\documentclass[a4paper,fleqn]{cas-dc}

\usepackage[numbers,sort&compress]{natbib}
\usepackage{amsmath}
\usepackage{multirow}
\usepackage{caption}
\def\tsc#1{\csdef{#1}{\textsc{\lowercase{#1}}\xspace}}
\tsc{WGM}
\tsc{QE}

\begin{document}
\let\WriteBookmarks\relax
\def\floatpagepagefraction{1}
\def\textpagefraction{.001}
\let\printorcid\relax



\title [mode = title]{Video Captioning with Aggregated Features Based on Dual Graphs and Gated Fusion}

%

















\author{Yutao Jin}

\author{Bin Liu}
\cormark[1]
\cortext[1]{Corresponding author}
\ead[1]{nj_liubin@163.com}
\author{Jing Wang}

\address{Nanjing Tech University,Nanjing,China}

\begin{abstract}
The application of video captioning models aims at translating the content of videos by using accurate natural language. Due to the complex nature inbetween object interaction in the video, the comprehensive understanding of spatio-temporal relations of objects remains a challenging task. Existing methods often fail in generating sufficient feature representations of video content. In this paper, we propose a video captioning model based on dual graphs and gated fusion: we adapt two types of graphs to generate feature representations of video content and utilize gated fusion to further understand these different levels of information. Using a dual-graphs model to generate appearance features and motion features respectively can utilize the content correlation in frames to generate various features from multiple perspectives. Among them, dual-graphs reasoning can enhance the content correlation in frame sequences to generate advanced semantic features; The gated fusion, on the other hand, aggregates the information in multiple feature representations for comprehensive video content understanding. The experiments conducted on worldly used datasets MSVD and MSR-VTT demonstrate state-of-the-art performance of our proposed approach.
\end{abstract}

\begin{keywords}
video captioning \sep graph neural network \sep gated fusion \sep
\end{keywords}

\maketitle

\section{Introduction}
With the continuous development of network communication technology and wide-spread application of intelligent mobile devices, a large amount of video data has been generated, which poses great challenges to video content understanding. Therefore, how to use deep learning technology to generate accurate descriptions for videos has become a hot topic. The video captioning model can convert videos into accurate natural language descriptions to reflect the content of videos as truly as possible. The video captioning model enables computers to have visual perception and semantic understanding capabilities, involving computer vision and natural language processing.

The current mainstream method for video captioning is the encoder-decoder architecture based on deep learning. Venugopalan et al. \cite{R1} use convolutional neural network to extract features from each sampled frame and use the average pooling strategy to generate an average frame feature vector representing the entire video content, which is input into LSTM (Long Short-Term Memory) to generate captions. Venugopalan et al. \cite{R2} proposed S2VT model, which utilizes LSTM to encode frame feature sequences to mine temporal information. The above methods utilize the frame level global information of videos but ignore the fine information of local object regions. In addition, simple encoding for features obtained by extractors can not explore the deeper semantic information of videos. Zhang et al. \cite{R3} consider the interaction between objects in each frame and construct an object relational graph to enhance object feature representations. The enhanced object features are combined with frame level appearance and motion features to represent video content. Xue and Zhou \cite{R4} calculate the cosine similarity between frame features and use frame features as nodes to construct a frame similarity graph (FSG) to generate high-level frame features. At the same time, the spatio-temporal relationships between objects are calculated and object features are as nodes to construct a Spatio-Temporal Aware Graph (STAG) to generate high-level object features. The methods elaborated above extract the features of each video from multiple perspectives but these features are only directly input into the Recurrent Neural Network for the generation of captions after simple unified processing, which fails to make full use of the information of each perspective of the feature, resulting in low utilization efficiency.

This paper proposes a video captioning method based on dual graphs and gated fusion, which fully utilizes the correlation in frame-frame and frame-object to generate features with richer semantic information and aggregates different features for better understanding of the video content. Specifically, 2D CNN, 3D CNN, and Faster-RCNN are used to extract appearance features, motion features, and object features respectively. In addition, two independent BiLSTMs are used to preprocess the appearance features and motion features respectively. Then, a Graph Attention Network (GAT) is constructed based on the preprocessed frame features for the interaction information between frames and a Frame Object Relational Graph (FORG) is constructed based on the preprocessed frame features and the extracted object features for the correlation between frames and objects. Dual-graphs reasoning can generate four feature sequences with deeper semantic information. Finally, we use multiple independent attention networks to obtain enhanced features and utilize the gated fusion method to fuse these various features as the input to the language LSTM to generate captions. Our method achieved great results on two benchmarks: MSVD and MSR-VTT.

The main contributions of this paper are as follows:
(1) Separating appearance features and motion features can generate more types of feature representations in subsequent stages, which helps to understand the content of videos from multiple perspectives.
(2) GAT and FORG are constructed to generate frame features with higher level semantic information, which fully explores the correlations in frame-frame and frame-object.
(3) Feature representations are fused based on multi-attention and gated fusion, which integrates information from multiple perspectives. The fused feature has a more comprehensive and accurate understanding of video content, enabling the generation of high quality captions.
\section{Related work}
\subsection{Video Captioning}
In the early stage of video captioning task development, fixed templates are used to generate captions. With the development of deep learning, especially the successful application of the deep network with encoder-decoder framework in tasks such as machine translation and image captioning, the video captioning model with encoder-decoder framework based on deep learning has become the current mainstream framework \cite{R5,R6,R7,R8,R9,R10,R11,R12,R13,R14,R15,R16}.

Yao et al. \cite{R5} utilize 3D CNN to extract temporal and spatial feature information from videos and use a temporal attention mechanism to assign corresponding attention weight to each 3D CNN frame feature to generate global context feature, which is input into the decoder LSTM to generate captions. Pei et al. \cite{R6} proposed MARN model, which stores and uses the memory visual context information of each word for multiple video scenes to provide important reference for the generation of captions. Zeng et al. \cite{R7} proposed VCRN model, which clusters the visual features of all videos in the dataset. Clustering centers implicitly represent visual commonsense concept or knowledge and combine with the visual features of the source videos to generate captions. Babariya and Tamaki \cite{R8} use YOLOv3 to extract several object regions in each frame and use the object region with the highest score to generate the object feature of that frame, which is encoded together with the frame feature to obtain richer visual feature representation. Zhang and Peng \cite{R9} proposed the OA-BTG model, which encodes object regions belonging to the same object entity to extract VLAD representations in forward and backward time order. A hierarchical attention mechanism is then used to generate the spatio-temporal features that integrate all object information, which is input into the decoder to generate captions. Tan et al. \cite{R10} proposed RMN model, designing three reasoning modules: locate module, relate module, and func module, as well as a module selector, to simulate human-level reasoning.

For the different types of features obtained, most of the existing methods choose to simply fuse them, which may not make full use of the semantic information in those features. This paper uses multi-attention and gated fusion to integrate the extracted features from multiple perspectives to help generate more detailed captions.

\subsection{Graph convolution neural network}
Graph convolution neural network is characterized by using the correlation between node features to obtain new features with rich interaction information. In recent years, graph convolutional neural network has been successfully applied in computer vision and natural language processing tasks such as image captioning \cite{R17,R18,R19}, video captioning \cite{R3,R20,R21,R22,R23,R24}, video question answering \cite{R25,R26}.

Yao et al. \cite{R19} design 11 spatial relationships between objects and construct the spatial object relational graph. Hua et al. \cite{R20} use a semantic segmentation tool to separate objects and surrounding scenes and extract relationship information between objects and surrounding scenes. An object scene relational graph is then constructed based on the extracted object-scene information and the semantic relationship between objects. Yan et al. \cite{R21} construct two kinds of object relational graphs by calculating the relative spatio-temporal position and similarity relationships between objects, as well as using object features as nodes. Xue and Zhou \cite{R22} calculate the cosine similarity between each frame feature and other frame features as edges and construct a frame similarity graph with frame features as nodes. Xiao et al. \cite{R23} use frame features as nodes and adopt a node-reduce strategy to update the graph, fusing information from multiple sequential frames into one node. Zhang et al. \cite{R3} respectively construct the partial object relational graph and the complete object relational graph based on the extracted object features. The above graphs determine the relationships between nodes in a self-defined way. Inspired by the self-attention mechanism in \cite{R27}, Veličković et al. \cite{R28} utilize this technique to determine the relationship between nodes and construct a new graph convolutional neural network called graph attention network. Li et al. \cite{R29} refer to graphs that use self-attention mechanism to calculate relationship between objects as implicit object relational graphs and also refer to graphs that extract relationship between objects by customized methods as explicit object relational graphs. Li et al. \cite{R26} use graph attention network with multi-head attention mechanism to obtain semantic relationships between objects and generate high-level object features.

Intuitively, visual features with deeper semantic information can be generated by combining the local information of objects with the global information of frames, which can represent the video content better. However, the above methods only focus on the interactions in object-object and frame-frame, which ignore the correlation between frames and objects. In this paper, we use FORG \cite{R24} and GAT to obtain different types of features to represent video content from multiple perspectives: FORG can obtain the correlation between frame features and object features to generate enhanced frame features containing object information; GAT is used to extract the correlation between frames to generate enhanced frame features containing the frame-frame interaction information.

\section{Methodology}
\begin{figure*}[h]
\centering
\includegraphics[width=18cm]{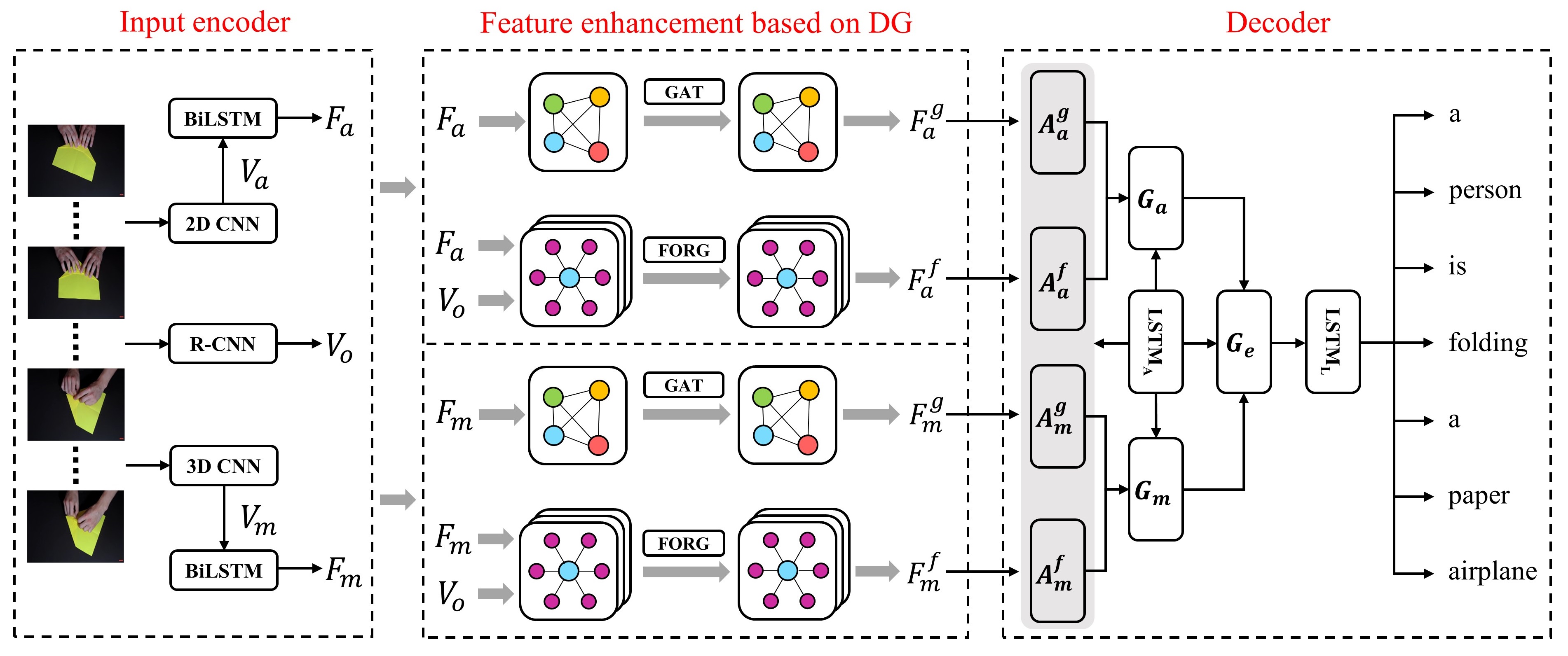}
\caption{The overview of our proposed video captioning model with aggregated features based on dual graphs and gated fusion.}
\label{F1}
\end{figure*}

The overall framework of our proposed video captioning model is shown in \hyperref[F1]{Fig. 1}, which is divided into three parts: Input encoder, Feature enhancement based on dual graphs and Decoder.

In the input encoder, the feature extractors are used to obtain corresponding appearance features $V_a$, motion features $V_m$ and object features $V_o$ for a video frame sequence respectively. Then, two independent BiLSTMs are used to mine temporal information in appearance features and motion features. In the feature enhancement stage, four types of frame features with high-level semantic information are generated by GAT and FORG based on the correlations in frame-frame and frame-object; In the decoder, we use multi-attention and gated fusion for four types of frame features to generate the fused feature with global semantic information, which is input into the language LSTM to generate captions.

\subsection{Input encoder}
\subsubsection{Feature extraction}
For a video frame sequence $X\!=\!\{ x_1,x_2,\dots,x_L \} $, 2D CNN and 3D CNN are respectively used to extract appearance features $V_a\!=\!\{ v_{a,1},v_{a,2},\dots,v_{a,L} \} $ and motion features $V_m\!=\!\{ v_{m,1},v_{m,2},\dots,v_{m,L} \} $. Faster-RCNN is used to extract object features $V_o\!=\!\{ v_{o,1}^1,\dots,v_{o,1}^N,\dots,v_{o,L}^1 ,\dots,v_{o,L}^N \} $, where $N$ denotes the number of object features in each frame. $R$ object features are extracted for a video in total, where $R\!=\!L \! \times \!  N$.

\subsubsection{Preprocessing based on BiLSTM}
Two independent BiLSTMs are used to mine temporal information in appearance and motion features to generate frame feature sequences $F_a$ and $F_m$, as shown in \hyperref[n1]{Eq. (1)} and \hyperref[n2]{Eq. (2)}.
\begin{flalign}
& F_a\!=\!BiLST \! M_a \! \left(V_a\right) & \label{n1} \\
& F_m\!=\!BiLST \! M_m \! \left(V_m\right) & \label{n2}
\end{flalign}
\subsection{Feature enhancement based on Dual-Graphs Reasoning}
\begin{figure*}[h]
\centering
\includegraphics[width=18cm]{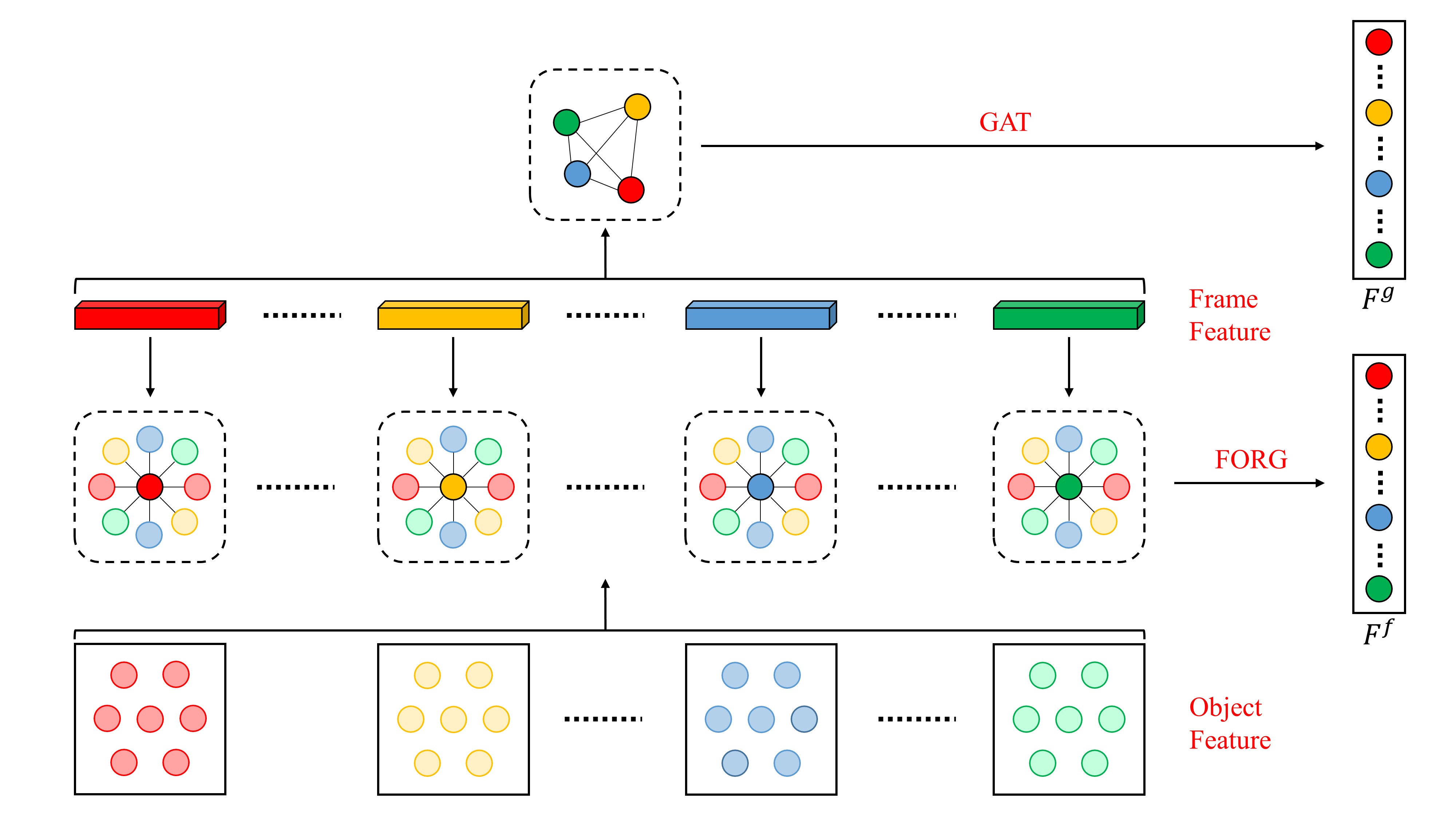}
\caption{Diagram for Dual-Graphs Reasoning.}
\label{F2}
\end{figure*}

When people watch videos, they often intend to gain an overall understanding of videos, instead of a separate understanding of each frame. Therefore, it is necessary to consider the content correlation between frames for generating captions. We use dual-graphs reasoning to obtain the content correlations between frames from four perspectives: two independent GATs are respectively constructed for the appearance feature sequence and motion feature sequence to extract two types of correlations between frames; Combined with the object information in each frame, two independent FORGs are respectively constructed for the appearance feature sequence and motion feature sequence to extract two types of correlations between frames and objects. The structure of dual-graphs reasoning is shown in \hyperref[F2]{Fig. 2}.
\subsubsection{Feature enhancement based on frame-frame interaction}
Most existing methods use relatively simple frame features processing methods, such as the average pooling and the temporal attention mechanism. In this paper, we use GAT to extract frame-frame correlation to construct a complete graph of frame features. The implicit relationship between frames is automatically learned and high-level frame feature sequences with rich frame-frame interaction information are generated based on GAT: Firstly, the weight $\alpha _{i,j}$ between the $i_{th}$ frame node and the $j_{th}$ neighbor node is calculated, as shown in \hyperref[n3]{Eq. (3)}; Then, the feature representation $f_i^g$ of the $i_{th}$ frame node enhanced by GAT is calculated, as shown in \hyperref[n4]{Eq. (4)}.
\begin{flalign}
& \alpha _{i,j} \! = \! \frac{\exp  \! \left ( LeakyReLU \! \left (  a^T \! \left [ W_gf_i;W_gf_j \right ] \right )  \right ) }{ {\textstyle \sum_{k=1}^{L}} \exp  \! \left ( LeakyReLU \! \left (  a^T \! \left [ W_gf_i;W_gf_k \right ] \right )  \right )} & \label{n3} \\
& f_i^g \!= \! \sigma  \! \left (  {\textstyle \sum_{j=1}^{L}}\alpha _{i,j}W_gf_j  \right ) & \label{n4}
\end{flalign}

Where $a^T \! \in \mathbb{R}^{2D_v^{'}}$,$W_g \! \in \mathbb{R}^{D_v^{'} \times D_v}$ denote learnable parameters and $f_i$, $f_j$ denotes the feature representation of the $i_{th}$ frame and the $j_{th}$ frame. $[;]$ represents the concatenation operation for features. $F_a$ and $F_m$ are enhanced by GAT to obtain new appearance feature $F_a^g \! =  \! \{ f_{a,i}^g \}$ and motion feature $F_m^g \! =  \! \{ f_{m,i}^g \}$:
\begin{flalign}
& F_a^g \! = \! G \! AT  \! \left( F_a  \right) & \label{n5} \\
& F_m^g \! = \! G \! AT  \! \left( F_m  \right) & \label{n6}
\end{flalign}

\subsubsection{Feature enhancement combined with object information}
Most of the existing methods only use object features to construct a relational graph to utilize the interaction information between objects. While it is conducive to make full use of the relationship between frame features and object features for the generation of more accurate captions. Combining the local information of objects with the global information of frames can obtain more comprehensive video information. Therefore, we use FORG to mine the correlation between frames and objects. Each frame feature and the object features of the entire video are used to construct a FORG and the correlation between each frame and all objects is used to enhance the frame features, which generate high-level semantic representations of the video: the kernel function is used to calculate the correlation between a frame feature and each object feature, as shown in \hyperref[n7]{Eq. (7)}; Then the correlation is used to calculate each enhanced frame feature $f_i^f$, as shown in \hyperref[n8]{Eq. (8)}.
\begin{flalign}
& F_{kernel} \! \left ( f_i,v_o^j \right )  \! = \! \psi   \! \left( f_i \right ) \! \phi \! \left( v_o^j  \right)^T & \label{n7} \\
& f_i^f  \! = \!  f_i+\sum_{j=1}^{R} F_{kernel} \! \left ( f_i,v_o^j \right ) W_fv_o^j & \label{n8}
\end{flalign}

where $\psi$ and $\phi$ both are linear layers followed by Tanh activation function. $W_f \!  \in \mathbb{R}^{D_p \times D_o}$ denotes learnable parameters. $v_o^j \!  \in \mathbb{R}^{D_o}$ denotes the $j_{th}$ object feature in the video and $f_i$ denotes the feature representation of the $i_{th}$ frame. Combined with object features, $F_a$ and $F_m$ are enhanced by FORGs to obtain new appearance features $F_a^f$ and motion feature $F_m^f$:
\begin{flalign}
& F_a^f \! = \! F \! O \! RG  \! \left( F_a,V_o  \right) & \label{n9} \\
& F_m^f \! = \! F \! O \! RG  \! \left( F_m,V_o  \right) & \label{n10}
\end{flalign}

Where $V_o$ represents the object features of the entire video.

\subsection{Decoder}
We follow the existing methods adopting the hierarchical LSTMs \cite{R3,R7,R24} as the decoder. The first level is the attention LSTM (denoted as $\mbox{LSTM}_A$ in \hyperref[n12]{Eq. (12)}) and its hidden state is regarded as the primary vocabulary representation. We also introduce a multi-attention module and a gated fusion module at this level: the multi-attention module can help to generate context features and the gated fusion module can help to aggregate context features. The second level is the language LSTM (denoted as $\mbox{LSTM}_L$ in \hyperref[n23]{Eq. (23)}), which receives the comprehensive feature with rich spatio-temporal information and the hidden state of $\mbox{LSTM}_A$. The output of $\mbox{LSTM}_L$ is used to generate fine vocabulary representation.
\subsubsection{Attention LSTM}
The last time step hidden state $h_{t-1}^L$ of $\mbox{LSTM}_L$, global average feature $\overline{F}$ (\hyperref[n11]{Eq. (11)}) and the last time step word embedding $w_{t-1}$ are concatenated as the input of $\mbox{LSTM}_A$. The output of $\mbox{LSTM}_A$ is the primary semantic information $h_t^A$ (\hyperref[n12]{Eq. (12)}) at the current time step.
\begin{flalign}
& \overline{F} \! = \! \frac{1}{L} \! \sum  \! \left ( \left [ F_a;F_m \right ] \right ) & \label{n11} \\
& h_t^A \! = \! LST \! M_A \! \left (\left [ h_{t-1}^L;\overline{F};w_{t-1} \right ],h_{t-1}^A\right ) & \label{n12}
\end{flalign}

Where $h_{t-1}^A$ denotes the last time step hidden state of $\mbox{LSTM}_A$.
\subsubsection{Multi-attention module}
We generate context features by using a mechanism similar to the self-attention model \cite{R24,R27} , as shown in \hyperref[n13]{Eq. (13)}. The hidden state of $\mbox{LSTM}_A$ is expanded as the query matrix $Q$ and the frame feature sequence is packaged as matrix $K$ and matrix $V$. Then we use four independent attention networks to generate four types of context features, as shown in \hyperref[n14]{Eq. (14--17)}.
\begin{flalign}
& A \! \left ( Q,K,V \right ) \! = \! so \! f \! tmax\left ( \frac{QK^T}{\sqrt{d_k}} \right )V & \label{n13} \\
& CG_t^a \! = \! A_a^g \! \left ( H_t^A,F_a^g,F_a^g \right ) & \label{n14} \\
& C \! F_t^a \! = \! A_a^f \! \left ( H_t^A,F_a^f,F_a^f \right ) & \label{n15} \\
& CG_t^m \! = \! A_m^g \! \left ( H_t^A,F_m^g,F_m^g \right ) & \label{n16} \\
& C \! F_t^m \! = \! A_m^f \! \left ( H_t^A,F_m^f,F_m^f \right ) & \label{n17}
\end{flalign}

Where $H_t^A$ denotes the matrix expanded by the hidden state $h_t^A$. $CG_t^a$ and $CG_t^m$ denote the context features generated by the appearance feature sequence and motion feature sequence enhanced by GAT, respectively. $C \! F_t^a$ and $C \! F_t^m$ denote the context features generated by the appearance feature sequence and motion feature sequence enhanced by FORG, respectively.

\subsubsection{Gated fusion module}
We adapt a gated fusion module \cite{R7} to hierarchically fuse the context features, which was generated by the multi-attention module. Different types of context features are fused based on the hidden state of $\mbox{LSTM}_A$, as shown in \hyperref[n18]{Eq. (18)} and \hyperref[n19]{Eq. (19)}. Two types of appearance context features and two types of motion context features respectively perform the first level fusion, as shown in \hyperref[n20]{Eq. (20)} and \hyperref[n21]{Eq. (21)}. Then, we use the second level fusion for the generated high-level appearance context feature $C_t^a$ and motion context feature $C_t^m$ to generate the final context feature $C_t^e$, as shown in \hyperref[n22]{Eq. (22)}.
\begin{flalign}
& \lambda \! = \! \sigma \! \left ( W_\lambda \! \left [X;Y;h_t^A \right ] \right ) & \label{n18} \\
& G \! \left( X,Y \right) \! = \! \lambda \odot  f \! \left(X\right)+ \! \left( 1-\lambda \right) \odot f \! \left( Y \right) & \label{n19} \\
& C_t^a \! = \! G_a \! \left(CG_t^a,C \! F_t^a\right) & \label{n20} \\
& C_t^m \! = \! G_m \! \left(CG_t^m,C \! F_t^m\right) & \label{n21} \\
& C_t^e \! = \! G_e \! \left(C_t^a,C_t^m\right) & \label{n22}
\end{flalign}
\subsubsection{Language LSTM}
The final context feature $C_t^e$ and the hidden state of $\mbox{LSTM}_A$ are concatenated as the input of $\mbox{LSTM}_L$, as shown in \hyperref[n23]{Eq. (23)}. Then the hidden state of $\mbox{LSTM}_L$ is used to generate the word probability distribution $p_t$, as shown in \hyperref[n24]{Eq. (24)}.
\begin{flalign}
& h_t^L \! = \! LST \! M_L \! \left (\left [ C_t^e;h_t^A \right ],h_{t-1}^L\right ) & \label{n23} \\
& p_t \! = \! so \! f \! tmax \! \left (M \! L \! P \! \left (h_t^L \right )\right ) & \label{n24}
\end{flalign}

Where $M \! L \! P$ is a two-layers perceptron with Tanh as activation function.

\subsubsection{Loss function}
We use cross-entropy loss to train the model:
\begin{flalign}
& L_{C \! E} \! = \! - \sum_{t=1}^{T} logP_t \! \left ( g_t^* \right ) & \label{n25}
\end{flalign}

Where $g_t^*$ is the ground-truth caption for a video and $T$ is the length of the caption.
\section{Experiments}
We conduct experiments on MSVD and MSR-VTT and use BLEU-4 \cite{R30}, METEOR \cite{R31}, ROUGE-L \cite{R32}, and CIDEr \cite{R33} as evaluation metrics.
\subsection{Datasets}
MSVD contains 1970 YouTube video clips. Each video clip is annotated with multiple languages. We use English annotations and each video clip has approximately 40 annotations. Following the previous work, we take 1200 video clips for training, 100 video clips for validation and 670 video clips for testing.

MSR-VTT contains 10000 video clips, divided into 20 categories. Each video clip has 20 English annotations. We split the dataset to 6513 video clips for training, 497 video clips for validation and 2990 video clips for testing.
\subsection{Data Preprocessing}
We follow the strategy illustrated in \cite{R10} to preprocess corpus and extract features: All captions are converted to lower case and punctuations are removed. Then, the length of each caption is fixed as 26, that is, the captions with more than 26 words are truncated and the captions with less than 26 words are padded with zeros. For MSVD, words that appear less than twice are removed and the vocabulary size is set to 7351; For MSR-VTT, words that appear less than 5 times are removed and the vocabulary size is set to 9732. 26 equally-spaced frames are obtained for each video. Inception ResNetV2 (IRV2) \cite{R34} is used as 2D CNN to extract appearance features and I3D \cite{R35} is used as 3D CNN to extract motion features. Faster-RCNN \cite{R36} is used to extract 36 object features in each frame.
\subsection{Training details}
We use the Adam optimizer to optimize the model. The initial learning rate is set to 0.0008. The batch size and the word embedding size are set to 128 and 300 for both datasets. The size of the hidden state of $\mbox{LSTM}_L$ is set to 1024 and 1536 for MSVD and MSR-VTT, respectively. The feature sizes in GAT and FORG are set to 1024 for both datasets. We use layer normalization in our model to accelerate convergence. We use beam search with size 5 to generate captions during inference.
\subsection{Comparison with State-of-the-Art}
\begin{table*}[h]
\caption{Result comparisons on MSVD and MSR-VTT. B@4 M, R, and C respectively represent BLEU-4, METROR, ROUGE-L, and CIDEr. The bold numbers represent the highest value of the metrics among all methods.}
\label{T1}
\begin{tabular*}{\linewidth}{@{}CCCCCCCCC@{}}
\toprule
\multirow{2}{*}{Models} & \multicolumn{4}{c}{MSVD} & \multicolumn{4}{c}{MSRVTT}\\ 
\cmidrule{2-5} \cmidrule{6-9}
&B@4&M&R&C&B@4&M&R&C\\
\midrule
OA-BTG \cite{R9}&56.9 &36.2 &-- &90.6 &41.4 &28.2 &-- &46.9 \\
RMN \cite{R10}&54.6 &36.5 &73.4 &94.4 &42.5 &28.4 &61.6 &49.6 \\
ORG-TRL \cite{R3}&54.3 &36.4 &73.9 &95.2 &43.6 &28.8 &62.1 &50.9 \\
TGN-RGN \cite{R23}&52.6 &36.3 &72.7 &89.6 &41.1 &28.6 &60.8 &48.8 \\
STAG-FSG \cite{R22}&58.6 &37.1 &73.0 &91.5 &43.3 &29.0 &\textbf{63.5} &49.5 \\
D-LSG \cite{R24}&\textbf{60.9} &37.6 &75.2 &100.8 &44.6 &28.8 &62.3 &51.2 \\
MFF-GATE \cite{R21}&56.4 &37.6 &73.4 &93.1 &43.9 &\textbf{30.2} &62.6 &50.8 \\
Ours&59.1 &\textbf{38.1} &\textbf{75.6} &\textbf{102.7} &\textbf{44.9} &29.0 &62.4 &\textbf{51.9} \\
\bottomrule
\end{tabular*}
\end{table*}
We compare some State-of-the-Art (SOTA) models with our proposed model. The SOTA models are divided into two categories: models that use object features but do not use graph neural network (GNN) and models that use both object features and GNN.

\subsubsection{Models that use object features but do not use GNN}
OA-BTG \cite{R9} uses a bidirectional temporal graph to process object features, which can well capture the spatio-temporal dynamic information of the same object entity. The model uses the object attention mechanism to aggregate the features of different object entities, which may ignore the interaction information between objects. In addition, the model also uses a bidirectional temporal graph and a temporal attention mechanism to process frame feature sequences, which takes the temporal information between frame features into account but ignore the interaction information. Object features in RMN \cite{R10} are fused with appearance features and motion features respectively, which consider the correlation between frames and objects but ignore the hidden correlation between frames.

\subsubsection{Models that use both object features and GNN}
ORG-TRL \cite{R3} constructs the object relational graph with GCN reasoning, which takes into account the interaction between object features but ignores the correlation between frame features; TGN-RGN \cite{R23} uses TGN to mine the interaction information between frame features and uses RGN to mine the interaction information between object features. STAG-FSG \cite{R22} uses FSG to mine the interaction information between frame features and uses STAG to mine the interaction information between object features. TGN-RGN and STAG-FSG do not consider the correlation between frame features and object features. D-LSG \cite{R24} uses latent proposal aggregation to aggregate the frame feature sequence fused with object information into several high-level feature representations. D-LSG considers the correlation between frame features and object features but ignores the correlation between frame features. MFF-GATE \cite{R21} uses a gate method to fuse appearance features and motion features and constructs Object Similarity Graph and Object G-IOU Graph to mine the interaction information between objects. The generated high-level object features and frame features are also fused based on the gate method.

The result comparisons are shown in \hyperref[T1]{Table 1}. For MSVD, our model achieves the best performance on METEOR, ROUGE-L, and CIDEr and the result on BLEU-4 is second only to D-LSG. For MSR-VTT, our model outperforms the SOTA models on BLEU-4 and CIDEr and the rest metrics also achieved comparable performance. It is crucial to have a comprehensive understanding of video content from multiple perspectives. Compared with the SOTA models, the reasons why our model achieves higher performance can be attributed to: (1) we simultaneously pay attention to the correlation in frame-frame and frame-object and extract multiple types of feature representations of video content; (2) The use of multi-attention and gated fusion can effectively utilize feature information from various perspectives.

\begin{table*}[h]
\caption{Ablation study on multiple feature representations.}
\label{T2}
\begin{tabular*}{\linewidth}{@{}CCCCCCCCCCCC@{}}
\toprule
\multirow{2}{*}{Models}& \multicolumn{3}{c}{Settings} & \multicolumn{4}{c}{MSVD} & \multicolumn{4}{c}{MSRVTT}\\ 
\cmidrule{2-4} \cmidrule{5-8} \cmidrule{9-12}
 &AMC&AMS&DG-GATE&B@4&M&R&C&B@4&M&R&C\\
\midrule
Baseline1 & \checkmark & &  &54.8 &36.3 &73.5 &95.1 &43.8 &28.3 &61.8 &49.7 \\
Baseline2 & & \checkmark &  &56.4 &36.7 &74.1 &98.1 &44.3 &28.3 &61.9 &50.3 \\
AMC-DG-GATE& \checkmark &  & \checkmark &58.0 &37.6 &74.2 &99.9 &44.8 &28.8 &62.1 &50.1 \\
AMS-DG-GATE& & \checkmark & \checkmark &59.1 &38.1 &75.6 &102.7 &44.9 &29.0 &62.4 &51.9 \\
\bottomrule
\end{tabular*}
\end{table*}

\begin{table*}[h]
\caption{Ablation study on feature enhancement based on dual graphs.}
\label{T3}
\begin{tabular*}{\linewidth}{@{}CCCCCCCCCCCC@{}}
\toprule
\multirow{2}{*}{Models}& \multicolumn{3}{c}{Settings} & \multicolumn{4}{c}{MSVD} & \multicolumn{4}{c}{MSRVTT}\\ 
\cmidrule{2-4} \cmidrule{5-8} \cmidrule{9-12}
 &GAT&FORG&GATE&B@4&M&R&C&B@4&M&R&C\\
\midrule
AMS-GAT-GATE & \checkmark & & \checkmark &58.0 &36.9 &74.1 &97.8 &44.1 &28.5 &61.9 &50.4 \\
AMS-FORG-GATE& & \checkmark & \checkmark &58.8 &37.7 &74.6 &100.5 &44.3 &28.8 &62.1 &50.3 \\
AMS-DG-GATE& \checkmark & \checkmark & \checkmark &59.1 &38.1 &75.6 &102.7 &44.9 &29.0 &62.4 &51.9 \\
\bottomrule
\end{tabular*}
\end{table*}

\begin{table*}[h]
\caption{Ablation study on gated fusion.}
\label{T4}
\begin{tabular*}{\linewidth}{@{}CCCCCCCCCCCC@{}}
\toprule
\multirow{2}{*}{Models}& \multicolumn{3}{c}{Settings} & \multicolumn{4}{c}{MSVD} & \multicolumn{4}{c}{MSRVTT}\\ 
\cmidrule{2-4} \cmidrule{5-8} \cmidrule{9-12}
 &GAT&FORG&GATE&B@4&M&R&C&B@4&M&R&C\\
\midrule
AMS-DG& \checkmark & \checkmark & &59.4 &37.3 &74.8 &97.0 &44.1 &28.3 &62.3 &48.9 \\
AMS-DG-GATE& \checkmark & \checkmark & \checkmark &59.1 &38.1 &75.6 &102.7 &44.9 &29.0 &62.4 &51.9 \\
\bottomrule
\end{tabular*}
\end{table*}
\subsection{Ablation Study}

To validate the effectiveness of our model (AMS-DG-GATE), ablation studies were conducted on MSVD and MSR-VTT to evaluate the effect on multiple feature representations, feature enhancement based on dual graphs and gated fusion, as shown in \hyperref[T2]{Table 2}, \hyperref[T3]{Table 3} and \hyperref[T4]{Table 4}, respectively.

Each ablation model is with various settings: (1) AMC denotes the concatenation of appearance features and motion features; (2) AMS denotes the separation of appearance features and motion features; (3) DG denotes the use of both GAT and FORG for feature enhancement; (4) GAT denotes the use of only GAT for feature enhancement; (5) FORG denotes the use of only FORG for feature enhancement; (6) GATE denotes the use of the gated fusion module.

\subsubsection{The effect of multiple feature representations}
The result comparisons are shown in \hyperref[T2]{Table 2}, where the Baseline1 model concatenates appearance features and motion features, while Baseline2 model separates appearance features and motion features. Baseline2 model achieves improvements on almost all of metrics for two datasets, which indicates that the separation of appearance features and motion features can achieve better results. This is because the information contained in appearance features and motion features has their own emphasis and separating them can help to fully utilize the unique semantic information in two types of frame features. AMC-DG-GATE model concatenates appearance features and motion features and then uses dual-graphs reasoning to obtain two types of feature representations. AMS-DG-GATE model separates appearance features and motion features and then obtains four types of feature representations by dual-graphs reasoning. AMS-DG-GATE model outperforms AMC-DG-GATE model on all evaluation metrics for both datasets, indicating that more types of feature representations are beneficial for mining the content information of videos and help to generate higher quality captions. The above two sets of comparisons both prove the effectiveness of multiple feature representations.
\subsubsection{The effect of feature enhancement based on dual graphs}
The result comparisons are shown in \hyperref[T3]{Table 3}, where AMS-GAT-GATE model only uses GAT reasoning on frame feature and AMS-FORG-GATE model introduces object feature for FORG reasoning. With the help of local fine information of the objects, AMS-FORG-GATE model performs better than AMS-GAT-GATE model. Compared with AMS-GAT-GATE model and AMS-FORG-GATE model, AMS-DG-GATE model has significant improvement on metrics for both datasets because dual-graphs reasoning can simultaneously focus on the correlations in frame-frame and frame-object and more sufficiently represent video content than single-graph reasoning. The above comparisons indicate that the effect of feature enhancement based on dual graphs is significantly better than that based on single graph, which proves the effectiveness of the dual-graphs method.

\subsubsection{The effect of gated fusion}
The result comparison is shown in \hyperref[T4]{Table 4}. AMS-DG model inputs the concatenation of multiple context features into $\mbox{LSTM}_L$ instead of using the gated fusion module. AMS-DG-GATE model inputs the aggregated feature generated by gated fusion into $\mbox{LSTM}_L$. Compared with AMS-DG model, our AMS-DG-GATE model improves on almost all metrics for both two datasets, especially CIDEr increasing by 5.7\% and 3\% on MSVD and MSR-VTT respectively. This is also because the gated fusion module can make full use of context features from multiple perspectives. The above comparison indicates that the use of gated fusion has great improvement on model performance, which proves the effectiveness of gated fusion.

\subsection{Qualitative Analysis}
\begin{figure*}[h]
\centering
\includegraphics[width=18cm]{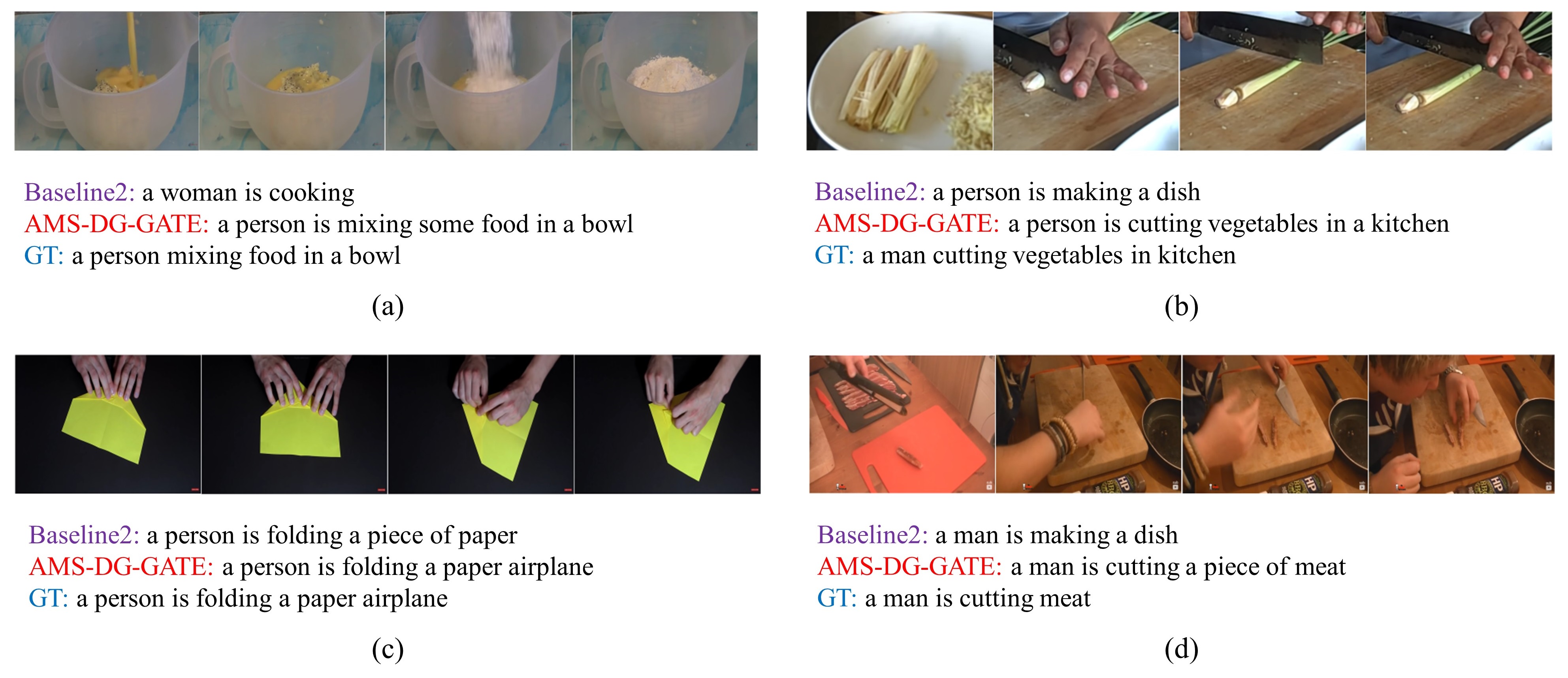}
\caption{The ground-truth captions and the captions generated by Baseline2 model and AMS-DG-GATE model.}
\label{F3}
\end{figure*}
We compare the captions generated by our proposed AMS-DG-GATE model and baseline2 model with the ground-truth captions (GT), as shown in \hyperref[F3]{Fig. 3}. It can be found that AMS-DG-GATE model can generate more accurate and detailed captions. 'a paper airplane' in example (c) and 'vegetables' in example (b) indicate that AMS-DG-GATE model can accurately recognize more detailed local object information due to the use of object feature. 'mixing some food' in example (a) and 'cutting a piece of meat' in example (d) reflect that AMS-DG-GATE model generates not only accurate words but also more detailed descriptions. ‘mixing’, 'cutting' and 'in a kitchen' indicate that AMS-DG-GATE model accurately focuses on motion and background information due to the interaction information captured by GAT and FORG and the comprehensive understanding of the feature representations from multiple perspectives generated by gated fusion.

\section{Conclusion}
This paper proposes a video captioning model based on dual graphs and gated fusion, which uses dual-graphs reasoning to generate multiple types of feature representations of video content and uses gated fusion to deeply understand these information from different perspectives. Firstly, the appearance features and motion features are respectively modeled by adopting dual-graphs reasoning to mine the correlations in frame-frame and frame-object to generate multiple types of feature representations. Then, the multi-attention mechanism are utilized to generate corresponding context features and then a gated fusion mechanism is used to fuse the multiple context features hierarchically. The generated fusion feature can more comprehensively and accurately represent video content, helping to generate higher quality captions. Our proposed model has achieved state-of-the-art performance on MSVD and MSR-VTT, proving the effectiveness of our model.

In future work, we will attempt to use different graph neural networks to handle the relationships between frame features and object features to the generate more types of feature representations. Futhermore, we will explore more efficient strategies in utilizing the miscellaneous feature representations.


\section*{CRediT authorship contribution statement}
\textbf{Yutao Jin:} Conceptualization, Methodology, Writing – original draft. \textbf{Bin Liu:} Supervision, Conceptualization, Methodology. \textbf{Jing Wang:} Validation, Writing – review $\&$ editing.

\section*{Declaration of Competing Interest}
The authors declare that they have no known competing financial interests or personal relationships that could have appeared to influence the work reported in this paper.

\section*{Acknowledgement}
This work was supported by National Natural Science Foundation of China under Grant 61672279.

\bibliographystyle{elsarticle-num-names}

\bibliography{cas-refs}

\bio{jinyutao}
Yutao Jin is now a post-graduate student in computer science at Nanjing Tech University. His research interests include deep learning, computer vision and natural language processing. 
\\
\\
\\
\\
\\
\endbio

\bio{liubin}
Bin Liu received the Ph.D. degree in management science and engineering from Southeast University, Nanjing, China. He is currently an Associate Professor with the School of Computer Science and Technology, Nanjing Tech University, Nanjing. His current research interests include deep learning and computer vision. 
\\
\\
\endbio

\bio{wangjing}
Jing Wang received her B.S. and M.S. degrees in electrical engineering from Nanjing Tech University, Nanjing, China, in 2007, and the Ph.D. degree in in electrical engineering from the Catholic University of America, Washington, DC, USA, in 2011. From 2007 to 2011, she was a Research Assistant with the Department of Electrical Engineering and Computer Science, the Catholic University of America. She is currently a Research Associate Professor at the Department of Electrical Engineering and Control Science, Nanjing Tech University. Her research interests include optimization theory, signal processing and communication theory in general, and wireless sensor localization in particular.
\endbio

\end{document}